\documentclass[sigconf, nonacm]{acmart}
\usepackage{fancyhdr}
\usepackage[ruled,vlined,linesnumbered]{algorithm2e}

\SetCommentSty{mycommfont}

\usepackage{multirow}

\usepackage{makecell}
\usepackage{color,soul}

\usepackage{pifont}

\usepackage{diagbox}
\usepackage{listings}

\usepackage{tikz}
\usepackage{amsmath}
\usepackage{amsfonts}
\usepackage{algorithmic}
\usepackage{graphicx}
\usepackage{textcomp}
\usepackage{xcolor}

\usepackage{color}

\newcommand\vldbdoi{XX.XX/XXX.XX}
\newcommand\vldbpages{XXX-XXX}
\newcommand\vldbvolume{14}
\newcommand\vldbissue{1}
\newcommand\vldbyear{2020}
\newcommand\vldbauthors{\authors}
\newcommand\vldbtitle{\shorttitle} 
\newcommand\vldbavailabilityurl{URL_TO_YOUR_ARTIFACTS}
\newcommand\vldbpagestyle{plain} 

\begin{document}
\title{Privacy Protection in Prosumer Energy Management Based on Federated
Learning}

\author{Yunfeng Li}
\affiliation{%
  \institution{the Department of Electronics and Information Engineering, Shenzhen University}
  \streetaddress{Nanshan District}
  \city{Shenzhen}
  \state{China}
  \postcode{518060}
}
\email{liyunfeng2019@email.szu.edu.cn}

\author{Xiaolin Li}
\affiliation{%
  \institution{the Department of Mathematics and Statistics, Shenzhen University}
  \streetaddress{Nanshan District}
  \city{Shenzhen}
  \country{China}
}
\email{2453043005@mails.szu.edu.cn}

\author{Gangqiang Li}
\affiliation{%
  \institution{the Department of Electronics and Information Engineering, Shenzhen University}
  \city{Shenzhen}
  \country{China}
}
\email{ligangqiang2017@email.szu.edu.cn}

\author{Zhitao Li}
\affiliation{%
  \institution{the Department of Electronics and Information Engineering, Shenzhen University}
  \city{Shenzhen}
  \country{China}
}
\email{2018092095@email.szu.edu.cn}



\begin{abstract}
With the booming development of prosumers, there is an urgent need for a prosumer energy management system to take full advantage of the flexibility of prosumers and take into account the interests of other parties. However, building such a system will undoubtedly reveal users' privacy. In this paper, by solving the  non-independent and identical distribution of data (Non-IID) problem in federated learning with federated cluster average(FedClusAvg) algorithm, prosumers' information can efficiently participate in the intelligent decision making of the system without revealing privacy. In the proposed FedClusAvg algorithm, each client performs cluster stratified sampling and multiple iterations. Then, the average weight of the parameters of the sub-server is determined according to the degree of deviation of the parameter from the average parameter. Finally, the sub-server multiple local iterations and updates, and then upload to the main server. The advantages of FedClusAvg algorithm are the following two parts. First, the accuracy of the model in the case of Non-IID is improved through the method of clustering and parameter weighted average. Second, local multiple iterations and three-tier framework can effectively reduce communication rounds. 
\end{abstract}

\maketitle

\pagestyle{\vldbpagestyle}
\begingroup\small\noindent\raggedright\textbf{PVLDB Reference Format:}\\
\vldbauthors. \vldbtitle. PVLDB, \vldbvolume(\vldbissue): \vldbpages, \vldbyear.\\
\href{https://doi.org/\vldbdoi}{doi:\vldbdoi}
\endgroup
\begingroup
\renewcommand\thefootnote{}\footnote{\noindent
This work is licensed under the Creative Commons BY-NC-ND 4.0 International License. Visit \url{https://creativecommons.org/licenses/by-nc-nd/4.0/} to view a copy of this license. For any use beyond those covered by this license, obtain permission by emailing \href{mailto:info@vldb.org}{info@vldb.org}. Copyright is held by the owner/author(s). Publication rights licensed to the VLDB Endowment. \\
\raggedright Proceedings of the VLDB Endowment, Vol. \vldbvolume, No. \vldbissue\ %
ISSN 2150-8097. \\
\href{https://doi.org/\vldbdoi}{doi:\vldbdoi} \\
}\addtocounter{footnote}{-1}\endgroup

\ifdefempty{\vldbavailabilityurl}{}{
\vspace{.3cm}
\begingroup\small\noindent\raggedright\textbf{PVLDB Artifact Availability:}\\
The source code, data, and/or other artifacts have been made available at \url{yunfengli.com}.
\endgroup
}

\section{Introduction}
\label{sec:introduction}

Distributed energy resources is a comprehensive energy utilization system distributed at the user end. It has the characteristics of reasonable utilization of energy efficiency, small loss, less pollution, flexible operation and good system economy. With the increasing popularity of distributed energy resources, traditional consumers are more and more likely to change to be prosumers. Different from the traditional consumers, prosumers not only participate in energy market transactions, but also provide decision support for energy management system\cite{xu2019distributed}. In order to give full play to the advantages of producers and take into account the interests of other parties\cite{ming2020prediction}, it is urgent to establish a prosumer energy management system \cite{injeti2020optimal}. However, construct such a prosumer energy management system  needs to use the user's information for comprehensive analysis and intelligent control to reduce the energy cost and improve the intelligent decision-making and planning ability of the existing energy management system, which will undoubtedly expose the privacy of users and may be  attacked through the data interaction process\cite{xu2020review}.

Moreover, in recent years, the issue of personal privacy has become more and more concerned. In 2016, the EU General Data Protection Regulation \cite{EU} required commercial companies to stop commercial companies from collecting, processing or exchanging user data without the permission of the corresponding user. Similarly, China and other countries are also introducing corresponding user data protection laws \cite{hourixin20200930}, to regulate the use of user data and protect user privacy\cite{9244122}. Fortunately, with the advent of federated learning, we can share data without violating users' privacy, and train efficient, intelligent and personal based energy management systems.

However, federated learning also faces many problems \cite{9084352}, such as data non-independent and identical distribution, communication\cite{9097889} and computing costs \cite{pmlr-v108-reisizadeh20a}, robustness, incentive mechanism \cite{9223632} and so on. At present, existing machine learning tasks usually default training data to follow independent and identically distributed (IID), and common algorithms such as neural networks and deep learning also use the assumption that data follows IID as part of their derivation. However, in some application scenarios of machine learning, such as information processing in energy management systems, medical image processing \cite{shin2016deep}, and federated learning, training data does not necessarily obey the assumption of independent and identical distribution. If simply treating all data as if it is the total data from IID, then the trained model will have systematic bias. Even if the training data is large enough, it cannot solve the problem caused by bias. Therefore, optimizing the algorithm under non-independent and identically distributed data in federated learning is of great significance to improving the accuracy of the model.

\subsection{Related Work}

Murat Dundar, Balaji Krishnapuram, Jinbo Bi, R. Bharat Rao, etc.\cite{dundar2007learning} used the reference point classifier to effectively improve the computational efficiency of the algorithm. H. Brendan McMahan, Eider Moore etc.\cite{mcmahan2017communication} proposed a deep network joint learning method based on iterative model averaging to solve the problem of non-independent and identical distribution of data (Non-IID)  data learning, and conducted a wide range of five different model structures and four data sets. Empirical evaluation.  The LoAdaBoost FedAvg algorithm proposed by Li Huang, Yifeng Yin etc.\cite{huang2020loadaboost} greatly improves the federated learning effect of Non-IID data by using a data sharing strategy. Wang Hao \cite{wanghao2015} proposed a Non-IID big data analysis method, which can better solve the problem of non-independent and non-identical distribution coupling relationships. Felix Sattler, Simon Wiedemann etc. \cite{FSSW} proposed a new data compression framework STC, which is specifically designed to meet the requirements of a joint learning environment. Wang Hao etc. \cite{HWZK} developed a control framework Favor driven by experience. In each round of federated learning, the client is intelligently selected to balance the parameter shift introduced by Non-IID data and accelerate the model convergence speed. Yuxuan Sun etc. \cite{YSSZ} proposed an online energy-aware dynamic work-hour scheduling strategy that introduces data redundancy into the system to process Non-IID data.

\vspace{5pt}

Li Huijuan \cite{lihuijuan2018} conducted research and analysis on the KNN classifier, and made improvements in decision rules and similarity measures around the deficiencies of the algorithm. Han Bing \cite{hanbing2019} designed the NI-PAM algorithm for numerical data by clustering and replacing the numerical data according to the attribute column according to the Euclidean distance to make the clustering effect better. Han Bing and Jiang He \cite{hanjiang2019} introduced numerical data non-independent and identically distributed calculation formulas for numerical data in the PAM algorithm, which greatly improved the clustering accuracy of the PAM algorithm. Pan Biying, Qiu Haihua, Zhang Jialun, etc. \cite{pqz2019} used the Minist and Cifar-10 data sets to study the stochastic gradient descent(SGD) algorithm of the federated machine learning model, and evaluated the accuracy of the shared model by controlling the number of clients participating in training and changing different data distributions.

\vspace{5pt}

Up to now, the method of studying federated learning Non-IID either adopts the method of sharing a small part of the data, which violates the original intention of protecting privacy and is not applicable in many scenarios; or the method of using a larger number of models, although effective but increasing communication costs. In addition, there is no related research results on the problem of non-independent and identical distribution of data in large clients, and there is no research on the problem that some parameters received by the server deviate from the average parameter in federated learning. Therefore, the research in this article is very meaningful.

\subsection{Contribution}

Our main contributions are:
\begin{enumerate}
	
	\item The simple and practical FedAvg algorithm is selected, and a certain degree of improvement is made based on this algorithm, and a new algorithm FedClusAvg algorithm is designed, which is more suitable for the direction of our research.
	
	\item In dealing with the problem of non-independent and identical distribution between clients, we use the degree of model parameter deviation from the average parameter to measure the weight of each client's update of the model parameters to achieve the effect of the model average and weaken the degree of parameter deviation.
	
	\item Multiple training of the model locally and the setting of a three-layer framework can effectively reduce the number of communication rounds.
\end{enumerate}
Moreover, our algorithm has been well verified on the public data set of the cardiovascular disease\footnote{https://www.kaggle.com/sulianova/cardiovascular-disease-dataset} on kaggle.

The rest of the paper is organized as follows. In Section \ref{l2}, we describe the federated learning related to this paper. In Section \ref{Intro1}, we describe our algorithm FedClusAvg in detail and set  a three-layer framework of FedClusAvg to reduce the number of communication rounds. In Section \ref{l4}, we introduce some experimental evaluation indicators to better evaluate the advantages of the algorithm. Simution results are given in Section \ref{l5} to confirm the effectiveness of FedClusAvg. We conclude this work in Section \ref{l6}.

\section{Federated Learning } \label{l2}

In recent years, artificial intelligence has developed rapidly, and a large number of machine learning-based applications such as computer vision, natural language processing, recommendation systems, speech recognition, etc. have all achieved great success. The success of these technologies is based on a large amount of data. However, in many fields,data cannot be arbitrarily obtained. Data privacy and security issues, and data island issues are the two main challenges hindering the further development of artificial intelligence.

\vspace{5pt}

Federated learning \cite{yang2019federated} is one of the techniques that can solve the above problems well. It can train machine learning models without having to concentrate all data on a central storage point. Similar to distributed learning, the difference is that federated learning adds the issue of privacy protection on the basis of distributed learning. The core idea of federated learning is to organize the nodes that have data sources, and each node is trained locally, and aggregated in the form of certain parameter information sharing of the model (rather than sharing private data) to obtain a global model. The performance of this global model is similar to the performance of training all the data together\cite{9153560}.

\begin{figure}[t!] \label{1}
	\centering
	\includegraphics[width=250pt]{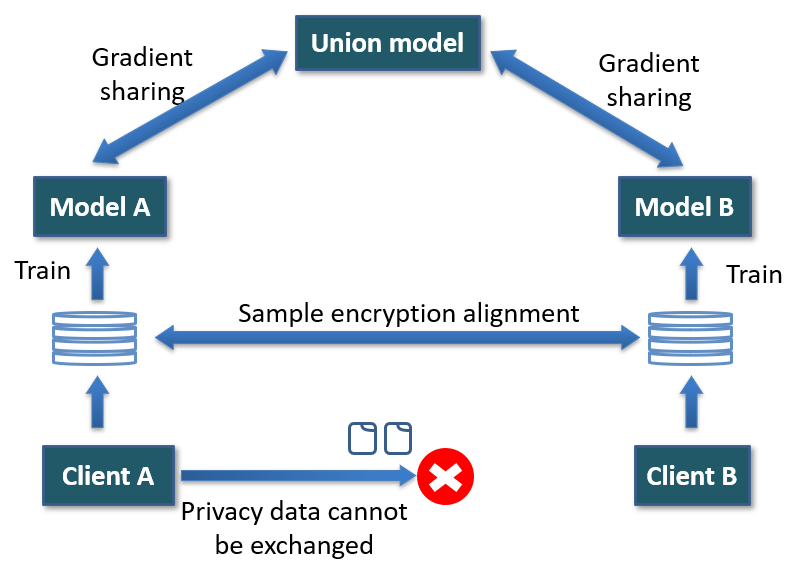}
	\caption{FedClusAvg algorithm}
	\label{fig:Fed}
\end{figure}

\vspace{5pt}
The algorithm in this paper can be applied in the following form: the privacy data of each client cannot be exchanged, each client sample is encrypted and aligned, and a model is trained locally, and then the model parameters are updated by gradient sharing to achieve the purpose of joint training model. Its form is shown in Figure \ref{1}.

\vspace{5pt}

As an innovative modeling learning mechanism, federated learning can establish a unified training model for multiple data sources without compromising the privacy and security of the data. Therefore, it will have a wide range of application scenarios in many fields. Take the energy management system as an example. With the vigorous development of my country’s consumer goods market, there is an urgent need to establish a user-oriented energy management system, such as the peer-to-peer approach, coordinated scheduling-based scheme and centralized control method. These strategies have some shortcomings, such as the lack of smart tools compatible with existing energy management systems, and the failure to improve user segmentation to increase the profitability of product consumers. At this time, it is a good way to solve the above problems by establishing a set of user data training model based on federated learning to participate in the auxiliary decision-making of the energy management system. It is expected that in the near future, federated learning will break down barriers between industries and establish a new model for data and information protection and sharing.

\section{Federated Cluster Average Algorithm}\label{Intro1}

Most successful applications of federated learning almost rely on the improved optimization of SGD, and the federated averaging algorithm (FedAvg) is also based on SGD to build a federated optimization algorithm. The FedAvg algorithm was first proposed by McMahan etc\cite{mcmahan2017communication}. It allows each client to iterate in parallel and synchronizes with the server after obtaining better parameters, which can effectively reduce the number of communications and make the model converge faster.

\vspace{5pt}

In federated learning, FedAvg algorithm can solve the problem of data Non-IID to a certain extent. However, the accuracy of FedAvg algorithm will be reduced when the imbalance of data feature distribution between clients is large. In this case, we can use the deviation degree of the parameters from each client to the service terminal relative to the average parameters to express the influence. When the parameter deviation of a client is large, we reduce the weight of the parameter, so that the algorithm can achieve higher accuracy in processing data Non-IID. At the same time, we notice that the data Non-IID problem in each client can not directly reflect the updating of model parameters. Therefore, we propose to split the clients with large distribution differences in the sample into smaller clients through sample clustering, and extract the training set by stratified sampling for multi-level iteration, so as to reduce the degree of non independent and identically distributed data, and then use weighted average method, the concept of gradient update can update the parameters of single client more accurately.

\vspace{5pt}

\subsection{Description of FedClusAvg Algorithm}

Based on the FedAvg algorithm, we constructed a new algorithm——federated cluster average algorithm (FedClusAvg ).

\vspace{5pt}

In FedClusAvg algorithm, we still divide federated learning into two parts: client and service terminal. The algorithm flow is shown in Figure \ref{fig:Fed}. Firstly, the personal client clusters the data according to the attribute value of the data, and extracts the training set of the client by stratified sampling; the client downloads the latest model parameter $\boldsymbol{w}_0$ from the service terminal, and then on the basis of the model parameter $\boldsymbol{w}_0$, useing the cluster sampling training set to train the model, performing multiple iterations in parallel, updating the model parameter $\boldsymbol{w}_k$ of the personal client $k$ , and then sending it to the service terminal. When the model parameter $\boldsymbol{w}_k$ returned by the personal client to the service terminal reaches a certain number, the service terminal will perform a summary update of the model parameters, generate the latest model parameter ${\boldsymbol{w}_0}^{'}$, and then perform a new round of model parameter update. The structure is shown in Figure \ref{fig:Fed}. The pseudo code of the FedClusAvg algorithm is shown in Table \ref{tab:FCA}.

\begin{figure}[t!]
	\centering
	\includegraphics[width=250pt]{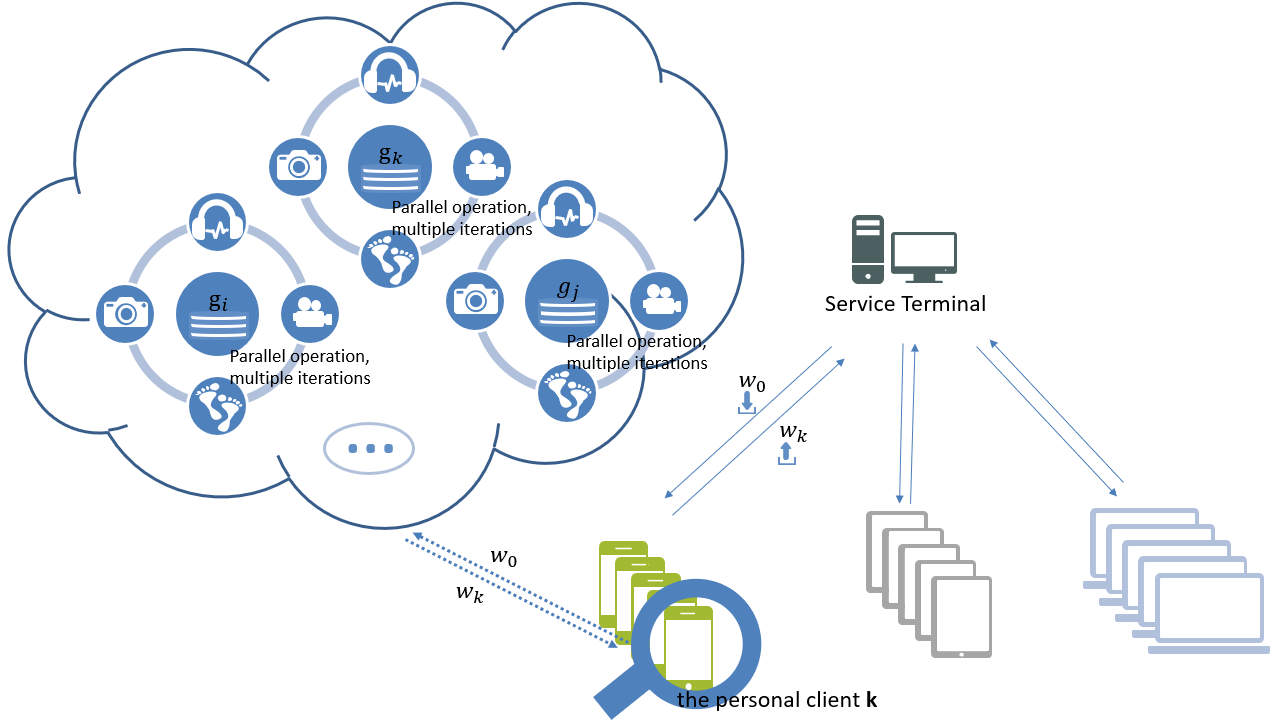}
	\caption{FedClusAvg algorithm}
	\label{fig:Fed}
\end{figure}

The FedClusAvg algorithm can be used for all objective functions with regard to a finite number of sample error accumulation functions:

\begin{equation}
\min_{\boldsymbol{w} \in R^{d}} f(\boldsymbol{w})
\centering
\label{eq1}
\end{equation}

\begin{equation}
f(\boldsymbol{w})=\frac{1}{n} \sum_{i=1}^{n} f_{i}(\boldsymbol{w})
\centering
\label{eq2}
\end{equation}

\begin{equation}
f_{i}(\boldsymbol{w})=l\left(x_{i}, y_{i} ; \boldsymbol{w}_{i}\right)
\centering
\label{eq3}
\end{equation}

In federated learning, assuming that there are $K$ individual clients participating in training, $P_k$ represents the local data set stored in the $k^{th}$ client, and the sample size is $n_k$, then the objective function is:

\begin{equation}
f(\boldsymbol{w}) =\sum_{k=1}^{K} \frac{n_{k}}{n} F_{k}(\boldsymbol{w})
\centering
\label{eq4}
\end{equation}

\begin{equation}
F_{k}(\boldsymbol{w}) =\frac{1}{n_{k}} \sum_{i \in P_{k}} f_{i}(\boldsymbol{w})
\centering
\label{eq5}
\end{equation}

\begin{equation}
f_{i}(\boldsymbol{w}) =l\left(x_{i}, y_{i} ; \boldsymbol{w}_{i}\right)
\centering
\label{eq6}
\end{equation}

\vspace{5pt}

\subsubsection{Personal Client}
In the personal client, in order to eliminate the Non-IID of the client sample $P_k$, this paper will cluster $P_k$, refining the data in the client into multiple small clients, and the client will become a small service terminal. In this case, the small client performs multiple rounds of parallel iteration, and the client performs a weighted average of the obtained parameters to obtain the latest model parameters, and then sends them to the service terminal.

\vspace{5pt}

When clustering client samples, the attribute values of each attribute of the sample need to be standardized to eliminate the influence caused by different units or large differences in measurement value ranges. In this paper, the sample in client $k$ is $x_{ki}=(x_{ki1},x_{ki2},\dots,x_{kip})^\mathrm{T},i=1,2,\dots,n_k$. $p$ is the number of sample attributes, and the standardized samples are:

\begin{equation}
x_{ki}^*=\frac{x_{ki}-\overline{x_{ki}}}{\sqrt{s_{ki}}}
\centering
\label{eq7}
\end{equation}

where $x_{ki}$ is the sample mean of the sample $x_{ki}$, and $s_{ki}$ is the sample variance of the sample $x_{ki}$.

\vspace{5pt}

Considering that the main purpose of client-side sample clustering is to eliminate the similarity between samples and the bias of proportion, this paper uses the system clustering method based on the longest distance method to cluster the samples. The steps are as follows. And the steps are as shown  in Table \ref{tab:SC}.
\begin{itemize}
	\item Select the threshold value $\theta \left(0<\theta<1\right)$.Take any sample as the first cluster center $Z_1$, such as $Z_1=x_1$.And the sample farthest from $Z_1$ is selected as the second cluster center $Z_2$.
	
	\item Looking for a new cluster center.
		\begin{enumerate}
			\item Calculate the distance between all other samples and the existing $J$ cluster centers:$D_{i1},D_{i2}, \ldots, D_{ij} \\ \left(i=1,2, \ldots, n_k\right)$.
			
			\item  if $D_s > \theta_{D_{12}}$,then $x_{s}$ is the new cluster center, otherwise stop looking for the cluster center, where
			
			\begin{equation}
			D_{s}=\max\left\{\min_{i}\left(D_{i1}, D_{i2}, \ldots, D_{ij}\right)\right\}
			\centering
			\label{eq8}
			\end{equation}
			
			\begin{equation}
			D_{12}=\left\|Z_{1}-Z_{2}\right\|=\sqrt{\left(Z_{1}-Z_{2}\right)^{2}}
			\centering
			\label{eq9}
			\end{equation}
		\end{enumerate}
	
	\item According to the principle of nearest neighbor, all samples are assigned to the nearest cluster center:
	\begin{center}
		if $D_{il}=\min_{j} D_{ij}$, then $x_i \in Z_l$
	\end{center}
\end{itemize}

\begin{table}[htbp]
	\caption{The pseudo code of the SpectralClust algorithm}
	\label{tab:SC}
	\begin{tabular}{lcl} 
		\toprule
		\textbf{SpectralClust algorithm}  Z is the set of cluster center $P_{ki} $ is the  $k$th \\client of the  $i$th sample. $J_{k}$ is cluster set of the large client $k$ and $J_{kj}$ \\ is the sample of $J_{k}$. \\ 
		\midrule
		Give a threshold  $\theta(0<\theta<1)$ \\
		$Z \leftarrow Z_{1}\quad (Z_{1} \in P_{k}$)\\
		$Z \leftarrow Z_{2}\quad (|Z_{2}-Z_{1}|\geq|Z_{i}-Z_{1}|,Z_{i}\in P_{k} \backslash Z_{1} $) \\
		for each $Z_{j} \in Z $ do\\
		\qquad for each $P_{ki} \in P_{k} \backslash Z $ do\\
		\qquad \qquad $\boldsymbol{D}_{i j} = \left\|\boldsymbol{p}_{ki}-\boldsymbol{Z}_{j}\right\|=\sqrt{\left(\boldsymbol{p}_{ki}-\boldsymbol{Z}_{j}\right)^{2}}$\\
		for each $p_{ki} \in P_{k} \backslash Z $ do\\
		\qquad $D_{i} =\max \left\{\mathop{min}\limits_{i}\left(D_{i 1}, D_{i 2}, \ldots, D_{i j}\right)\right\}$\\
		\qquad $D_{12} = \left\|\boldsymbol{Z}_{1}-\boldsymbol{Z}_{2}\right\|=\sqrt{\left(\boldsymbol{Z}_{1}-\boldsymbol{Z}_{2}\right)^{2}}$\\
		\qquad if $D_{i} \textgreater \theta D_{12}$\\ 
		\qquad \qquad $Z \leftarrow p_{ki}$\\ 
		for each $P_{ki} \in P_{k} \backslash Z $ do\\
		\qquad  $D_{i l} == \mathop{min}\limits_{j}D_{i j}$\\
		\qquad  $ J_{kl}\leftarrow P_{ki}$\\			
		\bottomrule
	\end{tabular} 
\end{table}

\vspace{5pt}

In order to simplify the algorithm process, this article sets the sample size of the client to be greater than 100 and the distance between the class centers of the optimal classification scheme must be greater than $50\%$ of the average distance between samples before classification, and the number of classifications is set to $\|\frac{n}{50}\|$.

\vspace{5pt}

After clustering, the client will become a small federated learning architecture. The algorithm flow in the client is roughly as follows: 

\begin{itemize}
	\item \textbf{Small clients obtained by clustering($j$ category, the sample size of each category is $n_{kj}$)}
	\begin{enumerate}
		\item Get the latest model parameter $\boldsymbol{w}_t$ from the personal client $k$; 
		\item In each small client, use the sample set $P_{kj}$ and parameter $\boldsymbol{w}_t$ in the small client to calculate the gradient $\boldsymbol{g}_{km}$ of the sample data set;
		\item Send the gradient $\boldsymbol{g}_{km}$ to the personal client $k$.
	\end{enumerate}
	
	\item \textbf{Personal Client $k$}
	\begin{enumerate}
		\item Obtain $\boldsymbol{g}_{k1},\boldsymbol{g}_{k2},\dots,\boldsymbol{g}_{kn_{kj}}$ from the small client, and calculate the weighted average:
		
		\begin{equation}
		\bar{\boldsymbol{g}}=\sum_{m=1}^{n_{kj}}\eta \boldsymbol{g}_{km}
		\centering
		\label{eq10}
		\end{equation}
		
		\begin{equation}
		\eta =\frac{\| \tilde{\boldsymbol{g}}-\boldsymbol{g}_{km}\|}{\sum_{m=1}^{n_{kj}} \|\tilde{\boldsymbol{g}}-\boldsymbol{g}_{km}\|}
		\centering
		\label{eq11}
		\end{equation}
		
		\begin{equation}
		\tilde{\boldsymbol{g}}=\sum_{m=1}^{n_{kj}}\frac{n_{kj}}{n_k}\boldsymbol{g}_{km}
		\centering
		\label{eq12}
		\end{equation}
		
		\item In each small client, use the sample set $P_{kj}$ and parameter $w_t$ in the small client to calculate the gradient $\boldsymbol{g}_{km}$ of the sample data set;
		\item Send the gradient $\boldsymbol{g}_{km}$ to the personal client $k$.
	\end{enumerate}
\end{itemize}

\subsubsection{Service Terminal}

On the server side, in order to minimize the impact of Non-IID data on the accuracy of the model, we propose to use the degree of parameter deviation from the average parameter as the weight of each client for the next round of model parameter update. When the degree of deviation is large, the corresponding drop reduces its weight, and when the degree of deviation is small, the weight increases.

\vspace{5pt}

The algorithm flow of the service terminal is roughly as follows:

\begin{enumerate}
	\item Obtain the updated parameters $\boldsymbol{w}_{(t+1)1},\boldsymbol{w}_{(t+1)2},\ldots,\\ \boldsymbol{w}_{(t+1)k}$ of each client in the current round from the personal client, and calculate the weighted average of the latest parameters value:
	
	\begin{equation}
	\boldsymbol{w}_{t+1}=\sum_{m=1}^K\eta \boldsymbol{w}_{(t+1)m}
	\centering
	\label{eq13}
	\end{equation}
	
	\begin{equation}
	\eta =\frac{\| \tilde{\boldsymbol{w}}-\boldsymbol{w}_{(t+1)m}\|}{\sum_{m=1}^{k} \|\tilde{\boldsymbol{w}}-\boldsymbol{w}_{(t+1)m}\|}
	\centering
	\label{eq14}
	\end{equation}
	
	\begin{equation}
	\quad \tilde{\boldsymbol{w}}=\sum_{m=1}^{K}\frac{n_k}{n} \boldsymbol{w}_{(t+1)m}
	\centering
	\label{eq15}
	\end{equation}

	\item Send the updated parameter $\boldsymbol{w}_{t+1}$ to each personal client for the next round of update.
\end{enumerate}

\begin{table}[h]
	\caption{The pseudo code of the FedClusAvg algorithm}
	\label{tab:FCA}
	\begin{tabular}{lcl} 
		\toprule
		\textbf{FedClusAvg algorithm}  The $K$ clients are indexed by $k$; $\alpha$ is the learning \\ rate; $ E$ is the number of iterations in each cluster set for each client;\\$ P_k$ is local dataset of the $k^{th}$ client used to update parameters; $ B$ is local\\ small batch size for client updates; $J_k$ is cluster set of the large client $k$.\\ 
		\midrule
		\textbf{Server executes: }\\
		\qquad initialize $w_0$\\
		\qquad for each round $t =1,2,\dots$ do\\
		\qquad \qquad for each client $k \in {1,2,\dots,K}$ in parallel do\\
		\vspace{5pt}
		\qquad \qquad \qquad $w_{t+1}^k \leftarrow$ ClientUpdate($k,w_t$)\\
		\vspace{5pt}
		\qquad \qquad $\tilde{w}=\sum_{m=1}^{K} \frac{n_k}{n} w_{t+1}^m$\\
		\vspace{5pt}
		\qquad \qquad $\eta^k \leftarrow \frac{\| \tilde{w}-w_{t+1}^k\|}{\sum_{k=1}^{K} \|\tilde{w}-w_{t+1}^k\|}$\\
		\vspace{10pt}
		\qquad \qquad $w_{t+1} \leftarrow \sum_{k=1}^K\eta^k w_{t+1}^k$\\
		
		\textbf{ClientUpdate($k,w$): }//Run on client $k$\\
		\qquad if $P_k>300$ do\\
		\qquad \qquad $J_k \leftarrow$ SpectralClust($P_k,J$)\\
		\qquad \qquad if $max(avg(dist(j\in J_k))<1.2avg(dist(p\in P_k))$\\
		\qquad \qquad \qquad $J_k \leftarrow J_k$\\
		\qquad \qquad \qquad else $J_k \leftarrow P_k$\\
		\qquad $B \leftarrow$ (split $J_k$ into batches of size $B$)\\
		\qquad for each class $j\in J_k $ do\\
		\qquad \qquad for each local epoch $i$ from 1 to E do\\
		\qquad \qquad \qquad for batch $b\in B$ do\\
		\qquad \qquad \qquad \qquad $w_j \leftarrow w_j-\alpha \nabla \ell(w_j,b)$\\
		\vspace{5pt}
		\qquad \qquad return $w_j$ to class $j$\\
		\vspace{5pt}
		\qquad $\bar{w} \leftarrow \sum_{j=1}^{J} \frac{n_j}{n_k} w_j$\\
		\vspace{5pt}
		\qquad $\eta_j \leftarrow \frac{\| \bar{w}-w_j\|}{\sum_{j=1}^{J} \|\bar{w}-w_j\|}$\\
		\vspace{5pt}
		\qquad $w_{t+1}^k \leftarrow \sum_{j=1}^J\eta^j w_j$\\
		\bottomrule
	\end{tabular} 
\end{table}

\subsection{FedClusAvg+ algorithm}

\begin{figure}[t!]
	\centering
	\includegraphics[width=230pt]{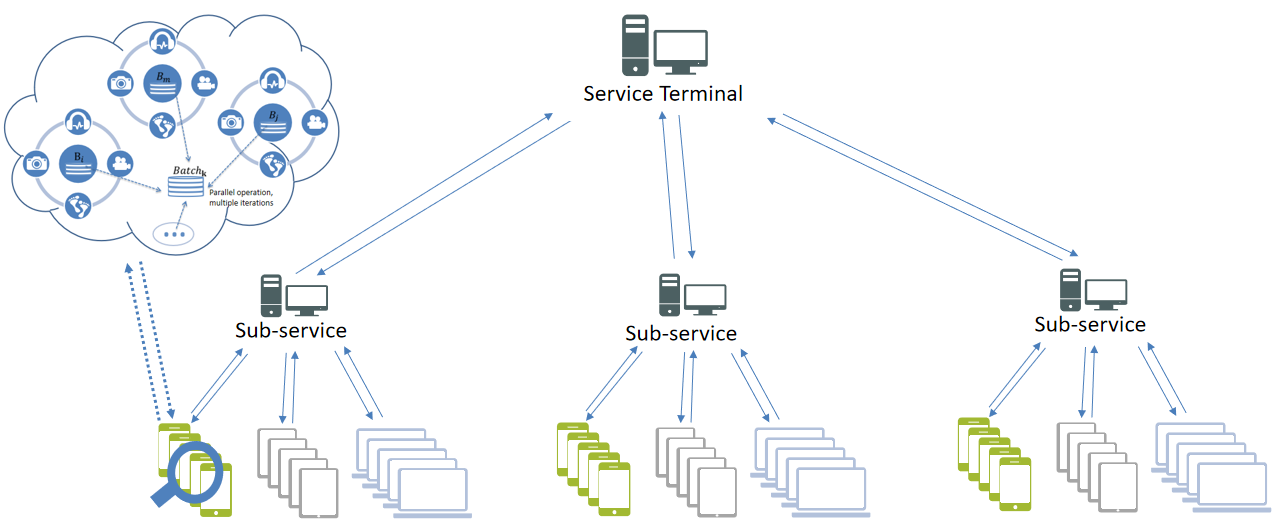}
	\caption{FedClusAvg+ algorithm}
	\label{fig:FCA1}
\end{figure}

Considering that in the practical application of Federated learning, servers are usually deployed in multiple locations. In FedClusAvg algorithm, the corresponding form of client and server is one-to-one, and the server achieves the goal of average gradient through communication. This paper uses an effective method to make the client interact with the second layer server many to one, and then the second layer server and the whole server take this as the starting point for transmission. Therefore, we change one server in FedClusAvg algorithm to multiple servers, and add a general central server. The structure is shown in Figure \ref{fig:FCA1}. And the pseudo code of the FedClusAvg+ algorithm is shown in Table \ref{tab:FCA1}.

\begin{table}[h]
	\caption{The pseudo code of the FedClusAvg+ algorithm}
	\label{tab:FCA1}
	\begin{tabular}{lcl} 
		\toprule
		\textbf{FedClusAvg+ algorithm}  The Q subservers are indexed by $q$; The $K$ clients \\of the $q^{th}$ subserver are indexed by $K_q$; $\alpha$ is the learning rate; $E$ is the \\ number of iterations in each cluster set for each client; $P_{kq}$ is local dataset \\ of the $k_q^{th}$ client used to update parameters; $B$ is local small batch size for \\ client updates; $J_kq$ is cluster set of the large client $k_q$.\\ 
		\midrule
		\textbf{Terminal Server executes: }\\
		\qquad initialize $w_0$\\
		\qquad for each round $t =1,2,\dots$ do\\
		\qquad \qquad for each subserver $q \in {1,2,\dots,Q}$ in parallel do\\
		\vspace{5pt}
		\qquad \qquad \qquad $w_{t+1}^q \leftarrow$ SubserverUpdate($Q,w_t$)\\
		\vspace{5pt}
		\qquad \qquad $\tilde{w}=\sum_{q=1}^{Q} \frac{n_q}{n} w_{t+1}^q$\\
		\vspace{5pt}
		\qquad \qquad $\eta^q \leftarrow \frac{\| \tilde{w}-w_{t+1}^q\|}{\sum_{q=1}^{Q} \|\tilde{w}-w_{t+1}^q\|}$\\
		\vspace{10pt}
		\qquad \qquad $w_{t+1} \leftarrow \sum_{q=1}^Q\eta^q w_{t+1}^q$\\
		
		\textbf{SubserverUpdate($Q,w_t$): }//Run on subserver $q$\\
		\qquad for each client $k \in {1,2,\dots,K_q}$ in parallel do\\
		\vspace{5pt}
		\qquad \qquad $w_{t+1}^k \leftarrow$ ClientUpdate($k_q,w_t$)\\
		\vspace{5pt}
		\qquad \qquad $\tilde{w^q}=\sum_{k=1}^{K_q} \frac{n_k}{n} w_{t+1}^k$\\
		\vspace{5pt}
		\qquad \qquad $\eta^{kq} \leftarrow \frac{\| \tilde{w^q}-w_{t+1}^k\|}{\sum_{k=1}^{K_q} \|\tilde{w^q}-w_{t+1}^k\|}$\\
		\vspace{10pt}
		\qquad \qquad $w_{t+1}^q \leftarrow \sum_{k=1}^{K_q}\eta^{kq} w_{t+1}^k$\\
		
		\textbf{ClientUpdate($k_q,w$): }//Run on client $k_q$\\
		\qquad if $P_{kq}>300$ do\\
		\qquad \qquad $J_{kq} \leftarrow$ SpectralClust($P_{kq},J$)\\
		\qquad \qquad if $max(avg(dist(j\in J_{kq}))<1.2avg(dist(p\in P_{kq}))$\\
		\qquad \qquad \qquad $J_{kq} \leftarrow J_{kq}$\\
		\qquad \qquad \qquad else $J_{kq} \leftarrow P_{kq}$\\
		\qquad $B \leftarrow$ (split $J_{kq}$ into batches of size $B$)\\
		\qquad for each class $j\in J_{kq} $ do\\
		\qquad \qquad for each local epoch $i$ from 1 to E do\\
		\qquad \qquad \qquad for batch $b\in B$ do\\
		\qquad \qquad \qquad \qquad $w_j \leftarrow w_j-\alpha \nabla \ell(w_j,b)$\\
		\vspace{5pt}
		\qquad \qquad return $w_j$ to class $j$\\
		\vspace{5pt}
		\qquad $\bar{w} \leftarrow \sum_{j=1}^{J} \frac{n_j}{n_k} w_j$\\
		\vspace{5pt}
		\qquad $\eta_j \leftarrow \frac{\| \bar{w}-w_j\|}{\sum_{j=1}^{J} \|\bar{w}-w_j\|}$\\
		\vspace{5pt}
		\qquad $w_{t+1}^k \leftarrow \sum_{j=1}^J\eta^j w_j$\\
		\bottomrule
	\end{tabular} 
\end{table}

\section{Indicator Description}\label{l4}

In order to comprehensively evaluate the overall predictive performance of the FedClusAvg algorithm, this article uses seven commonly used evaluation indicators in statistics, namely Accuracy, Precision, Recall, F-Measure, 
(receiver operating characteristic curve) ROC, Area Under Curve (AUC),  Kolmogorov-Smirnov (KS). Accuracy describes the proportion of all samples that are correctly predicted. Precison describes the proportion of samples predicted to be upright, which are actually positive. Recall describes the proportion of samples that are actually positive. F-Measure is equivalent to the harmonic mean of Precision and Recall.

Suppose the confusion matrix of the two classification model is as follows:
\begin{center}
\begin{tabular}{lcl} 
	\toprule
	&\textbf{Predicted positive}&\textbf{Predict negative} \\
		
	\midrule
	\textbf{Actually positive}&TP&EN\\
	\textbf{Actually negative}&FP&TN\\
	
	\bottomrule
\end{tabular} 
\end{center}

\vspace{8pt}
   
The calculation formula of the index is as follows:

\begin{equation}
Accuracy:P_a=\dfrac{T P+T N}{T P+F P+T N+F N}
\centering
\label{eq16}
\end{equation}

\begin{equation}
Precison:P_p=\dfrac{T P}{T P+F P}
\centering
\label{eq17}
\end{equation}

\begin{equation}
Recall:R=\dfrac{T P}{T P+F N}
\centering
\label{eq18}
\end{equation}

\begin{equation}
F-Measure:\dfrac{2}{F_{1}}=\dfrac{1}{P}+\dfrac{1}{R}
\centering
\label{eq19}
\end{equation}

\begin{equation}
F-Measure:F_{1}=\dfrac{2 T P}{2 T P+F P+F N}
\centering
\label{eq20}
\end{equation}

Before describing the Area Under roc Curve, we first give two definitions, namely TPR and FPR. TPR refers to Recall, while FPR refers to the proportion of negative samples that are incorrectly predicted to be positive. The two-class model returns a probability value, and multiple points about (FPR, TPR) can be obtained by adjusting the threshold, and the resulting curve is ROC. KS is Max (TPR-FPR). The larger the KS, the better the model can separate the interval of positive and negative samples, and the better the effect of the model.


\section{Case Study}\label{l5}

A cardiovascular disease public data set \footnote{https://www.kaggle.com/sulianova/cardiovascular-disease-dataset} on kaggle website is used to verify the feasibility and effectiveness of the proposed FedClusAvg algorithm. The dataset consists of 70 000 records of patients data,11 input features and 1 target tag.

\subsection{Experiment Setting}

\subsubsection{Data description}

The cardiovascular disease public data set is uploaded to kaggle by Svetlana ulianova, and all the data set values are collected during physical examination. The data set consists of 70, 000 physical examination records, including 11 input features and 1 target tag. The attribute features can be divided into four categories, namely, objective features, examination features and subjective features. Objective features are factual information, including age, height, weight and gender. Examination features are results of medical examination, including systolic blood pressure, diastolic blood pressure, cholesterol and glucose. Subjective features are information given by the patient, including whether they smoke, drink alcohol and exercise.

\subsubsection{Simulation settings}

According to Section \ref{Intro1}, the federal learning prediction model of cardiovascular disease is constructed based on FedClusAvg algorithm to predict whether physical examinees have cardiovascular disease. Through multiple tests and results analysis of the data set, this paper decided to add 11 features of the data set to the prediction model, that is, the prediction model has 11 input parameters, including age, height, weight, gender, systolic blood pressure, diastolic blood pressure, cholesterol, glucose, smoking, drinking and whether to exercise. The output of the prediction model is whether the physical examinee has cardiovascular disease.

In this simulation experiment, the data set was randomly divided into two parts. There are 60000 data records in the training set and 10000 data records in the test set. In order to reflect the phenomenon that federal learning data are not independent and identically distributed, this paper divides the training set into 100 customer data sets according to gender.

The initialization parameter of the prediction model is set to 0 and the learning rate is 0.01. The FedClusAvg algorithm and FedAvg algorithm are used as the federal learning training framework, and the logistic model is used for prediction. The above prediction models are programmed in the R x64 4.0.2 environment. The operating environment is Intel(R) Core (TM) i5-8250U CPU @1.60 GHz 1.80GHz, 8.00GB, Windows10. The corresponding parameters of this experiment are shown in Table 4.

\begin{table}[htbp]
	\caption{\label{tab:test}Parameter Settings}   \begin{tabular}{ccl}
		\toprule
		\bf{Parameter} & \bf{Parameter Value}	& \bf{Parameter Explanation} \\
		\midrule
		\multirow{2}{*}{k} & \multirow{2}{*}{100} & The number of clients divided by the \\
		~ & ~ & data set.\\
		
		\cmidrule(lr){1-3}
		Q & 5 &	The number of subservers.\\

		\cmidrule(lr){1-3}
		$\alpha$ & 0.01 &	Learning rate.\\
		\cmidrule(lr){1-3}
		\multirow{3}{*}{E} & \multirow{3}{*}{10} & The number of trainings performed\\
		~ & ~ & by each client on its local data set in \\
		~ & ~ & each round.\\
		\cmidrule(lr){1-3}
		M & 400 & Total update rounds on the server.\\
		\bottomrule
	\end{tabular}
\end{table}

\subsection{Prediction Results}

In the first part, two algorithms of FedClusAvg and FedAvg are used to perform federated learning on the data set, and then the test set is tested with the parameters of the training set iteration 400 times, and the accuracy of the parameters obtained by the two algorithms is compared. The accuracy of the change with the number of iterations is shown in Figure 4. The descriptive statistics table is shown in Table 5, which including accuracy, precision, recall and F-Measure. Each indicator are described in five parts namely min, quarter, median, mean and max. The curve of KS of two algorithms is shown in Figure 5, we can see that the value of KS of FedClusAvg algorithm is better than the value of FedAvg algorithm. 

\begin{figure}[t!]
	\centering
	\includegraphics[width=8cm]{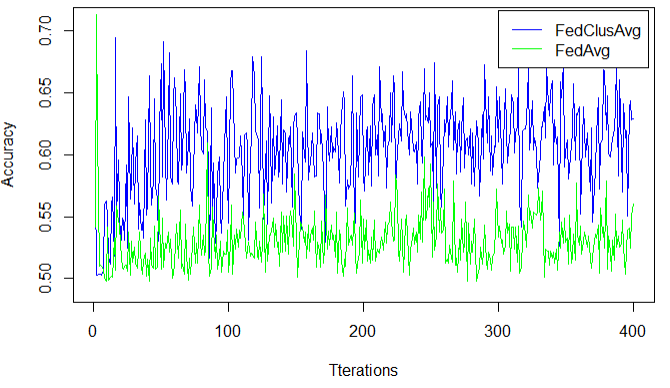}
	\caption{ the accuracy of  FedClusAvg and FedAvg}
\end{figure}

\begin{table}[t!]		
	\caption{\label{tab:test} evaluating indicator' datas  of  FedClusAvg and FedAvg}   \begin{tabular}{ccccccc}	
		\toprule
		& \bf{Min.} & \bf{1st Qu.} & \bf{Median} & \bf{Mean}  &\bf{Max.}\\
		\midrule
		FedClusAvg&\multirow{2}{*}{0.5026}&\multirow{2}{*}{0.5820}&\multirow{2}{*}{0.6091}&\multirow{2}{*}{0.6050}&\multirow{2}{*}{0.6947}\\
		Accuracy&~&~&~&~&~&~\\
		
		\cmidrule(lr){1-7}
		FedAvg&\multirow{2}{*}{0.4975}&\multirow{2}{*}{0.5153}&\multirow{2}{*}{0.5273}&\multirow{2}{*}{0.5303}&\multirow{2}{*}{0.7129}\\
		Accuracy&~&~&~&~&~&~\\
		
		\cmidrule(lr){1-7}
		FedClusAvg&\multirow{2}{*}{0.5026}&\multirow{2}{*}{0.5475}&\multirow{2}{*}{0.5666}&\multirow{2}{*}{0.5682}&\multirow{2}{*}{0.6857}\\
		Precison&~&~&~&~&~&~\\
		\cmidrule(lr){1-7}
		FedAvg&\multirow{2}{*}{0.5000}&\multirow{2}{*}{0.8260}&\multirow{2}{*}{0.8349}&\multirow{2}{*}{0.8315}&\multirow{2}{*}{0.8722}\\
		Precison&~&~&~&~&~&~\\
		\cmidrule(lr){1-7}
		FedClusAvg&\multirow{2}{*}{0.6852}&\multirow{2}{*}{0.9108}&\multirow{2}{*}{0.9451}&\multirow{2}{*}{0.9323}&\multirow{2}{*}{0.9976}\\
		Recall&~&~&~&~&~&~\\
		\cmidrule(lr){1-7}
		FedAvg&\multirow{2}{*}{0.00039}&\multirow{2}{*}{0.0443}&\multirow{2}{*}{0.073632}&\multirow{2}{*}{0.0819}&\multirow{2}{*}{0.6634}\\
		Recall&~&~&~&~&~&~\\
		\cmidrule(lr){1-7}
		FedClusAvg&\multirow{2}{*}{0.6684}&\multirow{2}{*}{0.6994}&\multirow{2}{*}{0.7081}&\multirow{2}{*}{0.7040}&\multirow{2}{*}{0.7213}\\
		$F_{1}$&~&~&~&~&~&~\\
		\cmidrule(lr){1-7}
		FedAvg&\multirow{2}{*}{0.00079}&\multirow{2}{*}{0.08415}&\multirow{2}{*}{0.135419}&\multirow{2}{*}{0.1438}&\multirow{2}{*}{0.6990}\\
		$F_{1}$&~&~&~&~&~&~\\
		\bottomrule
	\end{tabular}	
\end{table}		



\begin{figure}[t!]
	\centering
	\includegraphics[width=7.5cm]{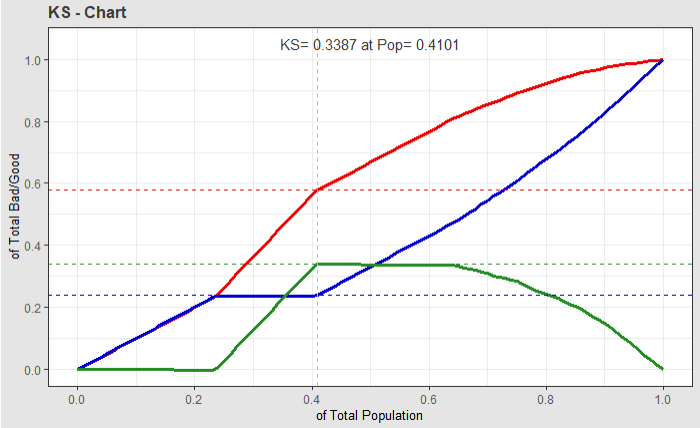}
	\includegraphics[width=7.5cm]{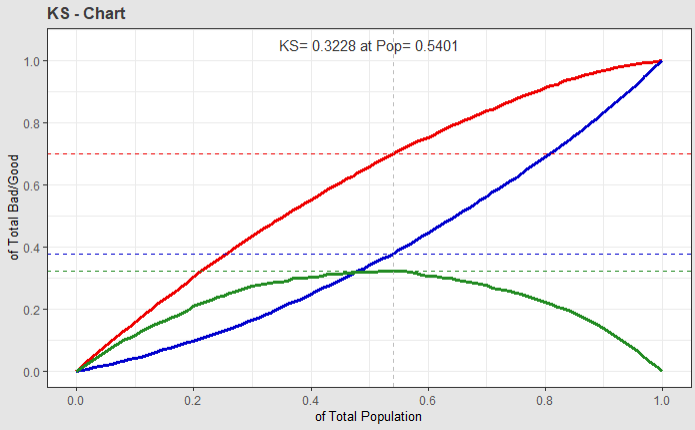}
	\caption{ The value of KS of FedClusAvg and FedAvg via the (Top) FedClusAvg algorithm and (Button) FedAvg algorithm }
\end{figure}

In the second part, except for the setting of Q value, the parameters of the two algorithms of FedClusAvg+ and FedAvg+ are the same as  the above algorithms. Finally, the accuracy of the change with the number of iterations is shown in Figure 6. The descriptive statistics table is shown in Table 6, which including accuracy, precision, recall and F-Measure. Each indicator are described in five parts namely min, quarter, median, mean and max. The curve of KS of two algorithms is shown in Figure 7, we can see that the value of KS of FedClusAvg+ algorithm is better than the value of KS of FedAvg+ algorithm. 

\begin{figure}[t!]
	\centering
	\includegraphics[width=8cm]{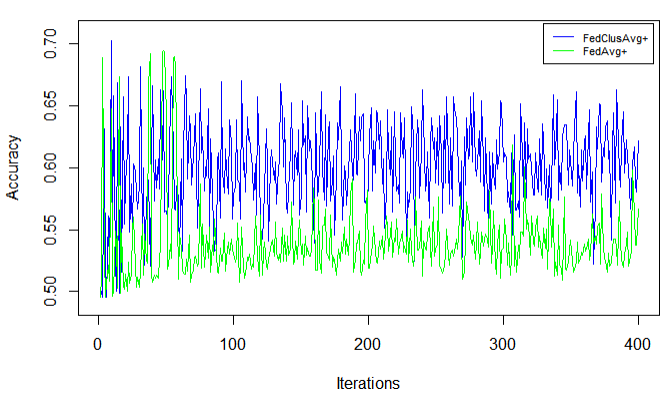}
	\caption{ the accuracy of  FedClusAvg+ and FedAvg+}
\end{figure}

\begin{table}[t!]		
	\caption{\label{tab:test} evaluating indicator' datas  of  FedClusAvg+ and FedAvg+}   \begin{tabular}{ccccccc}	
		\toprule
		& \bf{Min.} & \bf{1st Qu.} & \bf{Median} & \bf{Mean}  &\bf{Max.}\\
		\midrule
		FedClusAvg+&\multirow{2}{*}{0.4952}&\multirow{2}{*}{0.5777}&\multirow{2}{*}{0.6025}&\multirow{2}{*}{0.6008}&\multirow{2}{*}{0.7023}\\
		Accuracy&~&~&~&~&~&~\\
		
		\cmidrule(lr){1-7}
		FedAvg+&\multirow{2}{*}{0.4952}&\multirow{2}{*}{0.5218}&\multirow{2}{*}{0.5323}&\multirow{2}{*}{0.5388}&\multirow{2}{*}{0.6940}\\
		Accuracy&~&~&~&~&~&~\\
		
		\cmidrule(lr){1-7}
		FedClusAvg+&\multirow{2}{*}{0.4952}&\multirow{2}{*}{0.5414}&\multirow{2}{*}{0.5592}&\multirow{2}{*}{0.5634}&\multirow{2}{*}{0.7589}\\
		Precison&~&~&~&~&~&~\\
		\cmidrule(lr){1-7}
		FedAvg+&\multirow{2}{*}{0.4952}&\multirow{2}{*}{0.7565}&\multirow{2}{*}{0.7732}&\multirow{2}{*}{0.7436}&\multirow{2}{*}{0.8750}\\
		Precison&~&~&~&~&~&~\\
		\cmidrule(lr){1-7}
		FedClusAvg+&\multirow{2}{*}{0.5498}&\multirow{2}{*}{0.8945}&\multirow{2}{*}{0.9341}&\multirow{2}{*}{0.9160}&\multirow{2}{*}{0.9996}\\
		Recall&~&~&~&~&~&~\\
		\cmidrule(lr){1-7}
		FedAvg+&\multirow{2}{*}{0.00706}&\multirow{2}{*}{0.0594}&\multirow{2}{*}{0.0888}&\multirow{2}{*}{0.2115}&\multirow{2}{*}{0.9977}\\
		Recall&~&~&~&~&~&~\\
		\cmidrule(lr){1-7}
		FedClusAvg+&\multirow{2}{*}{0.6201}&\multirow{2}{*}{0.6903}&\multirow{2}{*}{0.6970}&\multirow{2}{*}{0.6946}&\multirow{2}{*}{0.7145}\\
		$F_{1}$&~&~&~&~&~&~\\
		\cmidrule(lr){1-7}
		FedAvg+&\multirow{2}{*}{0.01402 }&\multirow{2}{*}{0.1105 }&\multirow{2}{*}{0.15906 }&\multirow{2}{*}{0.23141 }&\multirow{2}{*}{0.7177 }\\
		$F_{1}$&~&~&~&~&~&~\\
		\bottomrule
	\end{tabular}	
\end{table}	



\begin{figure}[t!]
	\centering
	\includegraphics[width=7.5cm]{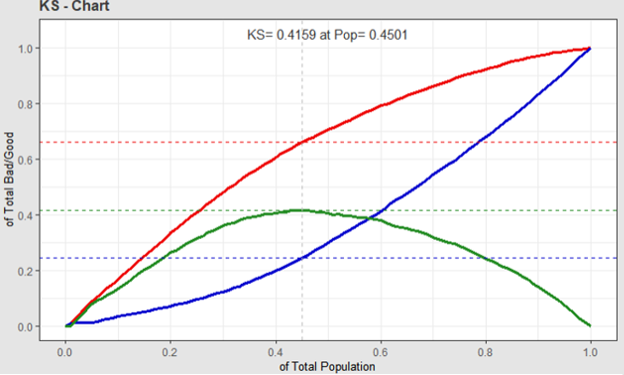}
	\includegraphics[width=7.5cm]{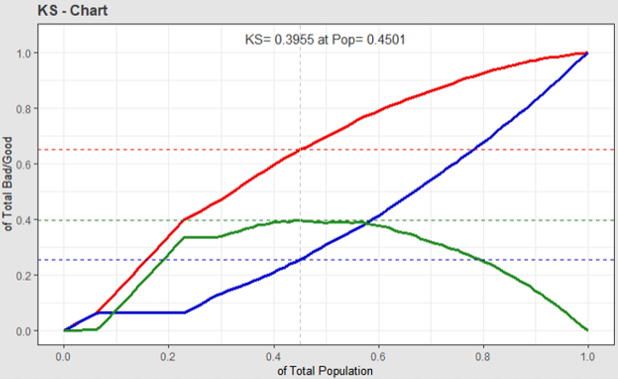}
	\caption{The value of KS of FedClusAvg+ and FedAvg+ via the (Top) FedClusAvg+ algorithm and (Button) FedAvg+ algorithm}
\end{figure}

Analysis of the above indicators shows that the algorithm in this paper is superior to the original algorithm in all evaluation indicators, whether it is FedClusAvg or FedClusAvg+. And the algorithm in this paper is more stable.

\subsection{application of FedClusAvg Algorithm}

\begin{figure}[t!] \label{energy}
	\centering
	\includegraphics[width=8cm]{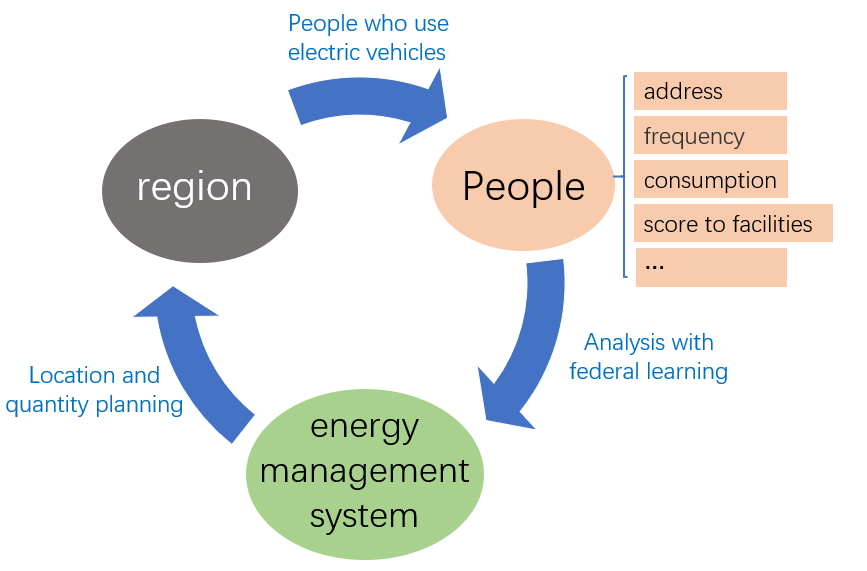}
	\caption{Location and quantity planning of charging point}
\end{figure}

In the energy consumption management system, the number and location planning of various energy production and transmission facilities\cite{huang2020multi}, such as charging piles, gas storage facilities, solar power stations, etc., are sometimes difficult to determine\cite{wang2020taxonomy}. Especially for areas with strong population mobility, if the population information is not fully mastered, it will lead to unreasonable distribution and waste of energy facilities. At this time, it is necessary to collect people's information, such as whether they are permanent residents, whether they use electric vehicles, the range and frequency of driving activities, energy consumption, etc., to make auxiliary decisions on the establishment of energy facilities address. Obviously, this will expose the user's privacy, which can be well protected through federated learning. However, due to the inherent defects of federated learning that data are not independent and distributed, the results of the model will be worse than that of non federated learning. The above example is described in Figure \ref{energy}. This paper proposes the FedClusAvg algorithm to solve this problem. The user information can be well involved in the construction of energy management system, so that the system can operate better.

\section{Conclusion}\label{l6}

This paper is mainly based on the FedclusAvg algorithm to solve the problem of Non-IID of data in federated learning. By improving and optimizing it, the algorithm results are more accurate and the model model can have a lower number of communications. The algorithm introduces the degree of deviation from the average parameter to correct the effect of the general gradient descent algorithm on the accuracy reduction when the client data distribution is large. At the same time, the client split method is used to reduce the internal data distribution difference of the client. The large impact on the model, the algorithm has achieved better results when the degree of Non-IID data is large.
%

\begin{acks}
\end{acks}


\bibliographystyle{ACM-Reference-Format}
\bibliography{sample}


\begin{thebibliography}{27}


\ifx \showCODEN    \undefined \def \showCODEN     #1{\unskip}     \fi
\ifx \showDOI      \undefined \def \showDOI       #1{#1}\fi
\ifx \showISBNx    \undefined \def \showISBNx     #1{\unskip}     \fi
\ifx \showISBNxiii \undefined \def \showISBNxiii  #1{\unskip}     \fi
\ifx \showISSN     \undefined \def \showISSN      #1{\unskip}     \fi
\ifx \showLCCN     \undefined \def \showLCCN      #1{\unskip}     \fi
\ifx \shownote     \undefined \def \shownote      #1{#1}          \fi
\ifx \showarticletitle \undefined \def \showarticletitle #1{#1}   \fi
\ifx \showURL      \undefined \def \showURL       {\relax}        \fi
\providecommand\bibfield[2]{#2}
\providecommand\bibinfo[2]{#2}
\providecommand\natexlab[1]{#1}
\providecommand\showeprint[2][]{arXiv:#2}

\bibitem[\protect\citeauthoryear{??}{EU}{[n.d.]}]%
        {EU}
 \bibinfo{year}{[n.d.]}\natexlab{}.
\newblock \showarticletitle{Regulation (EU) 2016/679 of the European Parliament and of the Council of 27 April 2016 on the protection of natural persons with regard to the processing of personal data and on the free movement of such data, and repealing Directive 95/46/EC (General Data Protection Regulation) (Text with EEA relevance)}.
\newblock \bibinfo{journal}{\emph{\underline{http://data.europa.eu/eli/reg/2016/679/oj}}} (\bibinfo{year}{[n.\,d.]}).
\newblock


\bibitem[\protect\citeauthoryear{{Aledhari}, {Razzak}, {Parizi}, and {Saeed}}{{Aledhari} et~al\mbox{.}}{2020}]%
        {9153560}
\bibfield{author}{\bibinfo{person}{M. {Aledhari}}, \bibinfo{person}{R. {Razzak}}, \bibinfo{person}{R.~M. {Parizi}}, {and} \bibinfo{person}{F. {Saeed}}.} \bibinfo{year}{2020}\natexlab{}.
\newblock \showarticletitle{Federated Learning: A Survey on Enabling Technologies, Protocols, and Applications}.
\newblock \bibinfo{journal}{\emph{IEEE Access}}  \bibinfo{volume}{8} (\bibinfo{year}{2020}), \bibinfo{pages}{140699--140725}.
\newblock
\urldef\tempurl%
\url{https://doi.org/10.1109/ACCESS.2020.3013541}
\showDOI{\tempurl}


\bibitem[\protect\citeauthoryear{Dundar, Krishnapuram, Bi, and Rao}{Dundar et~al\mbox{.}}{2007}]%
        {dundar2007learning}
\bibfield{author}{\bibinfo{person}{Murat Dundar}, \bibinfo{person}{Balaji Krishnapuram}, \bibinfo{person}{Jinbo Bi}, {and} \bibinfo{person}{R~Bharat Rao}.} \bibinfo{year}{2007}\natexlab{}.
\newblock \showarticletitle{Learning Classifiers When the Training Data Is Not IID.}. In \bibinfo{booktitle}{\emph{IJCAI}}. \bibinfo{pages}{756--761}.
\newblock


\bibitem[\protect\citeauthoryear{Han}{Han}{2019}]%
        {hanbing2019}
\bibfield{author}{\bibinfo{person}{Bing Han}.} \bibinfo{year}{2019}\natexlab{}.
\newblock \showarticletitle{Research on Non-IID k-medoids clustering Algorithm(in Chinese)}.
\newblock \bibinfo{journal}{\emph{Qilu University of Technology}} (\bibinfo{year}{2019}).
\newblock


\bibitem[\protect\citeauthoryear{Han and Jiang}{Han and Jiang}{2019}]%
        {hanjiang2019}
\bibfield{author}{\bibinfo{person}{Bing Han} {and} \bibinfo{person}{He Jiang}.} \bibinfo{year}{2019}\natexlab{}.
\newblock \showarticletitle{An improved PAM algorithm for numerical data under non-independent and identical distribution(in Chinese)}.
\newblock \bibinfo{journal}{\emph{Journal of Qilu University of Technology}} \bibinfo{volume}{33}, \bibinfo{number}{02} (\bibinfo{year}{2019}), \bibinfo{pages}{56--61}.
\newblock


\bibitem[\protect\citeauthoryear{Hou}{Hou}{9 30}]%
        {hourixin20200930}
\bibfield{author}{\bibinfo{person}{Rixin Hou}.} \bibinfo{year}{2020-09-30}\natexlab{}.
\newblock \showarticletitle{Scientific legislation to ensure data security(in Chinese)}.
\newblock \bibinfo{journal}{\emph{Learning times}} (\bibinfo{year}{2020-09-30}), \bibinfo{pages}{002}.
\newblock


\bibitem[\protect\citeauthoryear{Huang, Yin, Fu, Zhang, Deng, and Liu}{Huang et~al\mbox{.}}{2020b}]%
        {huang2020loadaboost}
\bibfield{author}{\bibinfo{person}{Li Huang}, \bibinfo{person}{Yifeng Yin}, \bibinfo{person}{Zeng Fu}, \bibinfo{person}{Shifa Zhang}, \bibinfo{person}{Hao Deng}, {and} \bibinfo{person}{Dianbo Liu}.} \bibinfo{year}{2020}\natexlab{b}.
\newblock \showarticletitle{LoAdaBoost: Loss-based AdaBoost federated machine learning with reduced computational complexity on IID and non-IID intensive care data}.
\newblock \bibinfo{journal}{\emph{Plos one}} \bibinfo{volume}{15}, \bibinfo{number}{4} (\bibinfo{year}{2020}), \bibinfo{pages}{e0230706}.
\newblock


\bibitem[\protect\citeauthoryear{Huang, Fang, and Deng}{Huang et~al\mbox{.}}{2020a}]%
        {huang2020multi}
\bibfield{author}{\bibinfo{person}{Zhao Huang}, \bibinfo{person}{Baling Fang}, {and} \bibinfo{person}{Jin Deng}.} \bibinfo{year}{2020}\natexlab{a}.
\newblock \showarticletitle{Multi-objective optimization strategy for distribution network considering V2G-enabled electric vehicles in building integrated energy system}.
\newblock \bibinfo{journal}{\emph{Protection and Control of Modern Power Systems}} \bibinfo{volume}{5}, \bibinfo{number}{1} (\bibinfo{year}{2020}), \bibinfo{pages}{7}.
\newblock


\bibitem[\protect\citeauthoryear{Injeti and Thunuguntla}{Injeti and Thunuguntla}{2020}]%
        {injeti2020optimal}
\bibfield{author}{\bibinfo{person}{Satish~Kumar Injeti} {and} \bibinfo{person}{Vinod~Kumar Thunuguntla}.} \bibinfo{year}{2020}\natexlab{}.
\newblock \showarticletitle{Optimal integration of DGs into radial distribution network in the presence of plug-in electric vehicles to minimize daily active power losses and to improve the voltage profile of the system using bio-inspired optimization algorithms}.
\newblock \bibinfo{journal}{\emph{Protection and Control of Modern Power Systems}} \bibinfo{volume}{5}, \bibinfo{number}{1} (\bibinfo{year}{2020}), \bibinfo{pages}{1--15}.
\newblock


\bibitem[\protect\citeauthoryear{{Kim}}{{Kim}}{2020}]%
        {9223632}
\bibfield{author}{\bibinfo{person}{S. {Kim}}.} \bibinfo{year}{2020}\natexlab{}.
\newblock \showarticletitle{Incentive Design and Differential Privacy Based Federated Learning: A Mechanism Design Perspective}.
\newblock \bibinfo{journal}{\emph{IEEE Access}}  \bibinfo{volume}{8} (\bibinfo{year}{2020}), \bibinfo{pages}{187317--187325}.
\newblock
\urldef\tempurl%
\url{https://doi.org/10.1109/ACCESS.2020.3030888}
\showDOI{\tempurl}


\bibitem[\protect\citeauthoryear{Li}{Li}{2018}]%
        {lihuijuan2018}
\bibfield{author}{\bibinfo{person}{Huijuan Li}.} \bibinfo{year}{2018}\natexlab{}.
\newblock \showarticletitle{Research on Non-IID KNN classification algorithm(in Chinese)}.
\newblock \bibinfo{journal}{\emph{Qilu University of Technology}} (\bibinfo{year}{2018}).
\newblock


\bibitem[\protect\citeauthoryear{{Li}, {Sahu}, {Talwalkar}, and {Smith}}{{Li} et~al\mbox{.}}{2020}]%
        {9084352}
\bibfield{author}{\bibinfo{person}{T. {Li}}, \bibinfo{person}{A.~K. {Sahu}}, \bibinfo{person}{A. {Talwalkar}}, {and} \bibinfo{person}{V. {Smith}}.} \bibinfo{year}{2020}\natexlab{}.
\newblock \showarticletitle{Federated Learning: Challenges, Methods, and Future Directions}.
\newblock \bibinfo{journal}{\emph{IEEE Signal Processing Magazine}} \bibinfo{volume}{37}, \bibinfo{number}{3} (\bibinfo{year}{2020}), \bibinfo{pages}{50--60}.
\newblock
\urldef\tempurl%
\url{https://doi.org/10.1109/MSP.2020.2975749}
\showDOI{\tempurl}


\bibitem[\protect\citeauthoryear{McMahan, Moore, Ramage, Hampson, and y~Arcas}{McMahan et~al\mbox{.}}{2017}]%
        {mcmahan2017communication}
\bibfield{author}{\bibinfo{person}{Brendan McMahan}, \bibinfo{person}{Eider Moore}, \bibinfo{person}{Daniel Ramage}, \bibinfo{person}{Seth Hampson}, {and} \bibinfo{person}{Blaise~Aguera y Arcas}.} \bibinfo{year}{2017}\natexlab{}.
\newblock \showarticletitle{Communication-efficient learning of deep networks from decentralized data}. In \bibinfo{booktitle}{\emph{Artificial Intelligence and Statistics}}. \bibinfo{pages}{1273--1282}.
\newblock


\bibitem[\protect\citeauthoryear{Ming, Xia, Lee, Adepoju, Shakkottai, and Xie}{Ming et~al\mbox{.}}{2020}]%
        {ming2020prediction}
\bibfield{author}{\bibinfo{person}{Hao Ming}, \bibinfo{person}{Bainan Xia}, \bibinfo{person}{Ki-Yeob Lee}, \bibinfo{person}{Adekunle Adepoju}, \bibinfo{person}{Srinivas Shakkottai}, {and} \bibinfo{person}{Le Xie}.} \bibinfo{year}{2020}\natexlab{}.
\newblock \showarticletitle{Prediction and assessment of demand response potential with coupon incentives in highly renewable power systems}.
\newblock \bibinfo{journal}{\emph{Protection and Control of Modern Power Systems}}  \bibinfo{volume}{5} (\bibinfo{year}{2020}), \bibinfo{pages}{1--14}.
\newblock


\bibitem[\protect\citeauthoryear{Pan, Qiu, and Zhang}{Pan et~al\mbox{.}}{2019}]%
        {pqz2019}
\bibfield{author}{\bibinfo{person}{Biying Pan}, \bibinfo{person}{Haihua Qiu}, {and} \bibinfo{person}{Jialun Zhang}.} \bibinfo{year}{2019}\natexlab{}.
\newblock \showarticletitle{Research on Federated machine learning technology with different data distribution(in Chinese)}. In \bibinfo{booktitle}{\emph{Proceedings of 5g Network Innovation Seminar (2019)}}. \bibinfo{pages}{6}.
\newblock


\bibitem[\protect\citeauthoryear{Reisizadeh, Mokhtari, Hassani, Jadbabaie, and Pedarsani}{Reisizadeh et~al\mbox{.}}{2020}]%
        {pmlr-v108-reisizadeh20a}
\bibfield{author}{\bibinfo{person}{Amirhossein Reisizadeh}, \bibinfo{person}{Aryan Mokhtari}, \bibinfo{person}{Hamed Hassani}, \bibinfo{person}{Ali Jadbabaie}, {and} \bibinfo{person}{Ramtin Pedarsani}.} \bibinfo{year}{2020}\natexlab{}.
\newblock \showarticletitle{FedPAQ: A Communication-Efficient Federated Learning Method with Periodic Averaging and Quantization}. In \bibinfo{booktitle}{\emph{Proceedings of the Twenty Third International Conference on Artificial Intelligence and Statistics}} \emph{(\bibinfo{series}{Proceedings of Machine Learning Research})}, \bibfield{editor}{\bibinfo{person}{Silvia Chiappa} {and} \bibinfo{person}{Roberto Calandra}} (Eds.), Vol.~\bibinfo{volume}{108}. \bibinfo{publisher}{PMLR}, \bibinfo{address}{Online}, \bibinfo{pages}{2021--2031}.
\newblock
\urldef\tempurl%
\url{http://proceedings.mlr.press/v108/reisizadeh20a.html}
\showURL{%
\tempurl}


\bibitem[\protect\citeauthoryear{{Sattler}, {Wiedemann}, {Müller}, and {Samek}}{{Sattler} et~al\mbox{.}}{2020}]%
        {FSSW}
\bibfield{author}{\bibinfo{person}{F. {Sattler}}, \bibinfo{person}{S. {Wiedemann}}, \bibinfo{person}{K.~R. {Müller}}, {and} \bibinfo{person}{W. {Samek}}.} \bibinfo{year}{2020}\natexlab{}.
\newblock \showarticletitle{Robust and Communication-Efficient Federated Learning From Non-i.i.d. Data}.
\newblock \bibinfo{journal}{\emph{IEEE Transactions on Neural Networks and Learning Systems}} \bibinfo{volume}{31}, \bibinfo{number}{9} (\bibinfo{year}{2020}), \bibinfo{pages}{3400--3413}.
\newblock
\urldef\tempurl%
\url{https://doi.org/10.1109/TNNLS.2019.2944481}
\showDOI{\tempurl}


\bibitem[\protect\citeauthoryear{Shin, Roth, Gao, Lu, Xu, Nogues, Yao, Mollura, and Summers}{Shin et~al\mbox{.}}{2016}]%
        {shin2016deep}
\bibfield{author}{\bibinfo{person}{Hoo-Chang Shin}, \bibinfo{person}{Holger~R Roth}, \bibinfo{person}{Mingchen Gao}, \bibinfo{person}{Le Lu}, \bibinfo{person}{Ziyue Xu}, \bibinfo{person}{Isabella Nogues}, \bibinfo{person}{Jianhua Yao}, \bibinfo{person}{Daniel Mollura}, {and} \bibinfo{person}{Ronald~M Summers}.} \bibinfo{year}{2016}\natexlab{}.
\newblock \showarticletitle{``{Deep convolutional neural networks for computer-aided detection: CNN architectures, dataset characteristics and transfer learning}''}.
\newblock \bibinfo{journal}{\emph{IEEE transactions on medical imaging}} \bibinfo{volume}{35}, \bibinfo{number}{5} (\bibinfo{year}{2016}), \bibinfo{pages}{1285--1298}.
\newblock


\bibitem[\protect\citeauthoryear{{Sun}, {Zhou}, and {Gündüz}}{{Sun} et~al\mbox{.}}{2020}]%
        {YSSZ}
\bibfield{author}{\bibinfo{person}{Y. {Sun}}, \bibinfo{person}{S. {Zhou}}, {and} \bibinfo{person}{D. {Gündüz}}.} \bibinfo{year}{2020}\natexlab{}.
\newblock \showarticletitle{Energy-Aware Analog Aggregation for Federated Learning with Redundant Data}. In \bibinfo{booktitle}{\emph{ICC 2020 - 2020 IEEE International Conference on Communications (ICC)}}. \bibinfo{pages}{1--7}.
\newblock
\urldef\tempurl%
\url{https://doi.org/10.1109/ICC40277.2020.9148853}
\showDOI{\tempurl}


\bibitem[\protect\citeauthoryear{Wang}{Wang}{2015}]%
        {wanghao2015}
\bibfield{author}{\bibinfo{person}{Hao Wang}.} \bibinfo{year}{2015}\natexlab{}.
\newblock \showarticletitle{Research of reliability based on the non-IIDness Learning(in Chinese)}.
\newblock \bibinfo{journal}{\emph{Harbin Engineering University}} (\bibinfo{year}{2015}).
\newblock


\bibitem[\protect\citeauthoryear{{Wang}, {Kaplan}, {Niu}, and {Li}}{{Wang} et~al\mbox{.}}{2020}]%
        {HWZK}
\bibfield{author}{\bibinfo{person}{H. {Wang}}, \bibinfo{person}{Z. {Kaplan}}, \bibinfo{person}{D. {Niu}}, {and} \bibinfo{person}{B. {Li}}.} \bibinfo{year}{2020}\natexlab{}.
\newblock \showarticletitle{Optimizing Federated Learning on Non-IID Data with Reinforcement Learning}.
\newblock \bibinfo{journal}{\emph{IEEE INFOCOM 2020 - IEEE Conference on Computer Communications, Toronto, ON, Canada}} (\bibinfo{year}{2020}), \bibinfo{pages}{1698--1707}.
\newblock
\urldef\tempurl%
\url{https://doi.org/10.1109/INFOCOM41043.2020.9155494}
\showDOI{\tempurl}


\bibitem[\protect\citeauthoryear{Wang, Liu, Zhou, Li, Cao, Voropai, and Barakhtenko}{Wang et~al\mbox{.}}{2020}]%
        {wang2020taxonomy}
\bibfield{author}{\bibinfo{person}{Huaizhi Wang}, \bibinfo{person}{Yangyang Liu}, \bibinfo{person}{Bin Zhou}, \bibinfo{person}{Canbing Li}, \bibinfo{person}{Guangzhong Cao}, \bibinfo{person}{Nikolai Voropai}, {and} \bibinfo{person}{Evgeny Barakhtenko}.} \bibinfo{year}{2020}\natexlab{}.
\newblock \showarticletitle{Taxonomy research of artificial intelligence for deterministic solar power forecasting}.
\newblock \bibinfo{journal}{\emph{Energy Conversion and Management}}  \bibinfo{volume}{214} (\bibinfo{year}{2020}), \bibinfo{pages}{112909}.
\newblock


\bibitem[\protect\citeauthoryear{Xu, Wu, Zhou, Li, Bai, and Huang}{Xu et~al\mbox{.}}{2019}]%
        {xu2019distributed}
\bibfield{author}{\bibinfo{person}{Da Xu}, \bibinfo{person}{Qiuwei Wu}, \bibinfo{person}{Bin Zhou}, \bibinfo{person}{Canbing Li}, \bibinfo{person}{Li Bai}, {and} \bibinfo{person}{Sheng Huang}.} \bibinfo{year}{2019}\natexlab{}.
\newblock \showarticletitle{Distributed Multi-Energy Operation of Coupled Electricity, Heating, and Natural Gas Networks}.
\newblock \bibinfo{journal}{\emph{IEEE Transactions on Sustainable Energy}} \bibinfo{volume}{11}, \bibinfo{number}{4} (\bibinfo{year}{2019}), \bibinfo{pages}{2457--2469}.
\newblock


\bibitem[\protect\citeauthoryear{Xu}{Xu}{2020}]%
        {xu2020review}
\bibfield{author}{\bibinfo{person}{Yan Xu}.} \bibinfo{year}{2020}\natexlab{}.
\newblock \showarticletitle{A review of cyber security risks of power systems: from static to dynamic false data attacks}.
\newblock \bibinfo{journal}{\emph{Protection and Control of Modern Power Systems}} \bibinfo{volume}{5}, \bibinfo{number}{1} (\bibinfo{year}{2020}), \bibinfo{pages}{1--12}.
\newblock


\bibitem[\protect\citeauthoryear{{Yan}, {Liu}, {Jin}, {Qian}, {Zhang}, and {Lu}}{{Yan} et~al\mbox{.}}{2020}]%
        {9097889}
\bibfield{author}{\bibinfo{person}{Y. {Yan}}, \bibinfo{person}{S. {Liu}}, \bibinfo{person}{Y. {Jin}}, \bibinfo{person}{Z. {Qian}}, \bibinfo{person}{S. {Zhang}}, {and} \bibinfo{person}{S. {Lu}}.} \bibinfo{year}{2020}\natexlab{}.
\newblock \showarticletitle{Risk Minimization Against Transmission Failures of Federated Learning in Mobile Edge Networks}.
\newblock \bibinfo{journal}{\emph{IEEE Access}}  \bibinfo{volume}{8} (\bibinfo{year}{2020}), \bibinfo{pages}{98205--98217}.
\newblock
\showISSN{2169-3536}
\urldef\tempurl%
\url{https://doi.org/10.1109/ACCESS.2020.2996307}
\showDOI{\tempurl}


\bibitem[\protect\citeauthoryear{Yang, Liu, Chen, and Tong}{Yang et~al\mbox{.}}{2019}]%
        {yang2019federated}
\bibfield{author}{\bibinfo{person}{Qiang Yang}, \bibinfo{person}{Yang Liu}, \bibinfo{person}{Tianjian Chen}, {and} \bibinfo{person}{Yongxin Tong}.} \bibinfo{year}{2019}\natexlab{}.
\newblock \showarticletitle{``{Federated machine learning: Concept and applications}''}.
\newblock \bibinfo{journal}{\emph{ACM Transactions on Intelligent Systems and Technology (TIST)}} \bibinfo{volume}{10}, \bibinfo{number}{2} (\bibinfo{year}{2019}), \bibinfo{pages}{1--19}.
\newblock


\bibitem[\protect\citeauthoryear{{Zhu}, {Mong Goh}, and {Ng}}{{Zhu} et~al\mbox{.}}{2020}]%
        {9244122}
\bibfield{author}{\bibinfo{person}{H. {Zhu}}, \bibinfo{person}{R.~S. {Mong Goh}}, {and} \bibinfo{person}{W.~K. {Ng}}.} \bibinfo{year}{2020}\natexlab{}.
\newblock \showarticletitle{Privacy-Preserving Weighted Federated Learning Within the Secret Sharing Framework}.
\newblock \bibinfo{journal}{\emph{IEEE Access}}  \bibinfo{volume}{8} (\bibinfo{year}{2020}), \bibinfo{pages}{198275--198284}.
\newblock
\showISSN{2169-3536}
\urldef\tempurl%
\url{https://doi.org/10.1109/ACCESS.2020.3034602}
\showDOI{\tempurl}


\end{thebibliography}

\end{document}